\documentclass[runningheads]{llncs}
\usepackage[T1]{fontenc}
\usepackage{graphicx}
\usepackage{booktabs}
\usepackage{multirow}
\usepackage{amssymb}
\usepackage[misc]{ifsym}
\usepackage{amsmath}

\usepackage{mwe}
\usepackage{subcaption}
\usepackage{multirow}

\begin{document}

\title{Streaming Hierarchical Inference with Tabular Foundation Models}

\titlerunning{Streaming Hierarchical Inference with Tabular Foundation Models}

\author{Vítor Crista\textsuperscript{*} \and Afonso Lourenço\textsuperscript{*} \and Diogo Martinho \and Goreti Marreiros}

\authorrunning{Crista et al.}

\institute{GECAD, Polytechnic of Porto, Portugal  \textsuperscript{*}Equal contribution}

\maketitle              

\begin{abstract}
Tabular Foundation Models (TFMs) have recently demonstrated strong predictive performance through in-context learning, but their deployment in high-throughput data streams remains challenging due to communication overhead and latency. We propose \textit{HINT}, a hierarchical inference framework that combines edge-based retrieval with cloud-based TFM inference. A graph-based approximate nearest neighbor memory maintained over a sliding window provides local predictions and uncertainty estimates, allowing confident samples to be processed locally while uncertain instances are selectively offloaded, together with their retrieved context, to a cloud-hosted TFM. The framework exposes an offloading threshold and a neighborhood retrieval policy that can be varied to balance predictive performance and communication cost. Experiments show \textit{HINT} consistently identifies favorable trade-offs.
\end{abstract}

\section{Introduction}

The Internet of Things (IoT) connects a vast network of physical devices, generating massive, high-speed data streams \cite{cossu2026practical}. For tabular stream learning (SL), ensembles of incremental decision trees (IDTs) have long been state-of-the-art \cite{neves2025online,lourencco2026ihomer+}. They rely on statistical bounds to determine node splits and handle concept drift through subtree replacement. As shallow learners, IDTs achieve fast online convergence due to their limited number of trainable parameters \cite{lourencco2026dfdt}. However, their representational capacity is constrained by single-view feature splits, loss of plasticity from locally optimal decisions, catastrophic forgetting in class-conditional estimators, and an inability to capture complex dependencies \cite{neves2026pitfalls}. Alternatively, continuous sequential architectures, such as LSTMs and RNNs, are well suited for streams, with structurally evolving frameworks offering promising solutions to catastrophic forgetting under evolving distributions \cite{giannini2026dynamic}. However, the inductive biases of these DL architectures assume structures which offer little advantage for the irregular patterns typical of tabular data.

\begin{figure}
    \centering
    \includegraphics[width=\textwidth]{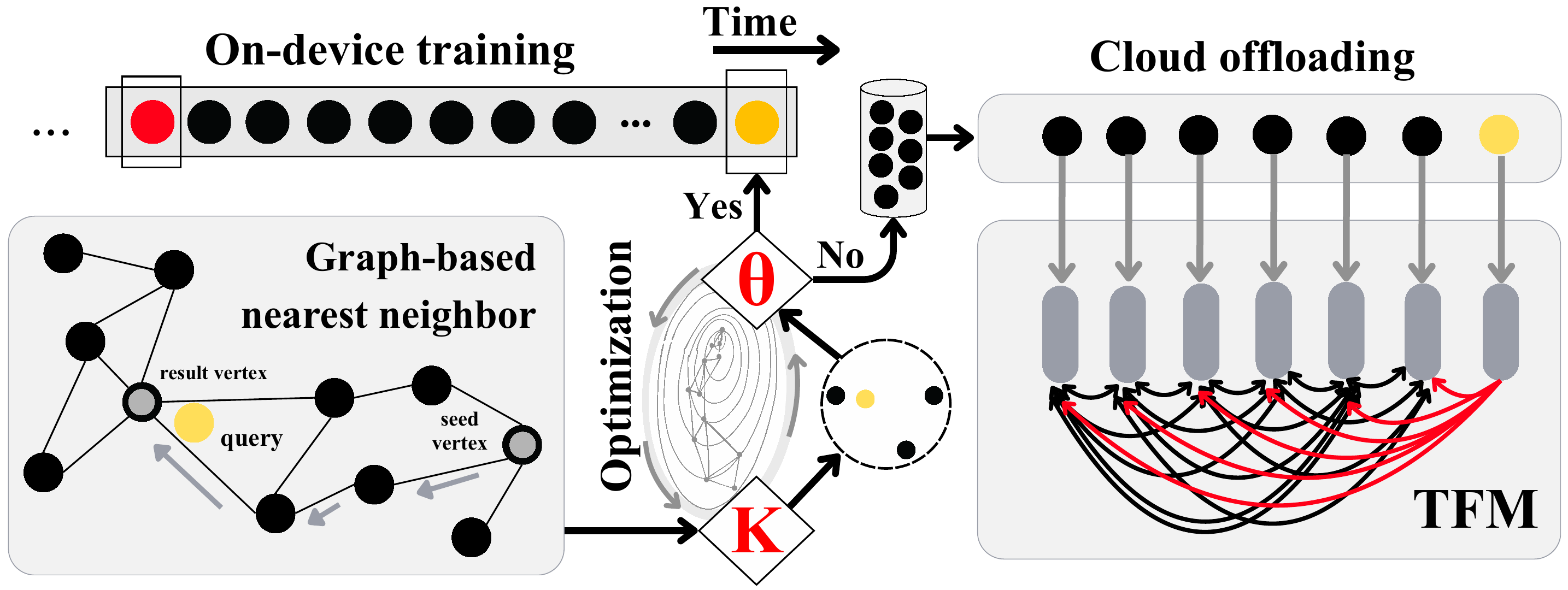}
    \caption{\textbf{Streaming hierarchical inference} defined by \textbf{(1)} confidence threshold $\theta$ for offloading uncertain predictions; and \textbf{(2)} number of neighbors $k$. Both varied to balance misclassifications and communication costs.}
    \label{fig:vamos}
\end{figure}

In contrast, the recent success of tabular foundation models (TFMs) \cite{grinsztajn2025tabpfn} has introduced emergent capabilities such as few-shot in-context learning, opening a new direction for streaming continual learning. Unlike traditional tabular models, TFMs perform classification without fine-tuning \cite{grinsztajn2025tabpfn}, adapting to unseen datasets in a single forward pass by treating training examples as contextual information \cite{lourencco2026context}. From a stream learning perspective, TFMs enable immediate processing of incoming data. However, their deployment must account not only for the learning paradigm but also for system constraints such as network capacity and execution location \cite{pauperio2025explainable}. Machine learning systems are generally divided into edge solutions (TinyML up to 100KB RAM, MobileML up to 8GB), optimized for resource-constrained devices, and server-based CloudML solutions, which prioritize accuracy over computational efficiency \cite{warden2019tinyml}.

As of the state-of-the-art, TFMs belong to the CloudML category, relying on remote servers for in-context inference. While powerful, this setup can become inefficient under high-throughput, real-time data streams due to communication overhead, and latency. To mitigate these, edge–cloud collaboration strategies enable selective communication between local devices and centralized models. Building on the two-stage nature of in-context stream mining of (1) local context retrieval followed by (2) in-context inference \cite{feuer2023scaling,lourencco2025bridging,lourencco2026context}, we propose \textit{HINT: Hierarchical Inference with Neighborhood Transfer}, where the local retriever acts as a kNN-based predictor providing uncertainty estimates to decide whether to predict locally or offload to a TFM in the cloud \cite{behera2025exploring}. The inference configuration is governed by the confidence threshold $\theta$ for offloading uncertain predictions; and the number of retrieved neighbors $k$, with an objective function that jointly minimizes the costs of misclassification and offloading costs. At the server, we use TabPFN, a prior-data fitted transformer, as the TFM \cite{grinsztajn2025tabpfn}. At the edge, we use a sliding window graph-based approximate nearest neighbor method tailored for sublinear nearest neighbor queries in streaming settings, using NN-Descent refinement to support efficient insertion, deletion, and search operations within bounded memory \cite{mastelini2024swinn}. We make three contributions: (i) formulate TFM-based stream prediction as an edge-cloud hierarchical inference problem; (ii) propose a graph-based sliding-window retrieval mechanism that acts both as local predictor and context selector; (iii) empirically characterize the accuracy–communication Pareto frontier across benchmarks.

\section{Related work}

TFMs enable in-context learning for tabular data, performing classification without task-specific training by conditioning on contextual examples at inference time \cite{lourencco2026context}. While this makes them appealing for stream learning \cite{cossu2026practical}, their deployment is often limited to cloud environments, creating a tension between predictive performance and the resource constraints of edge devices \cite{lourencco2025device}.

Most existing approaches address this through model-centric compression. For example, TabPFN has explored model distillation into compact surrogate models such as MLPs and tree ensembles \cite{grinsztajn2025tabpfn}, while alternative approaches employ hypernetworks to generate task-specific predictors in a single forward pass \cite{bonet2024hyperfast}. Other work investigates early-stopping strategies that dynamically determine whether to terminate in-context processing after intermediate Transformer layers \cite{kuken2025early}. Although effective, these methods focus on reducing model complexity rather than reducing the amount of information required for inference.

In contrast, we adopt a data-centric perspective. Since TFMs adapt their effective representational capacity according to the size and structure of the provided context \cite{lourencco2026context}, the central challenge becomes maintaining a compact yet informative memory from which relevant contextual examples can be efficiently retrieved. This recovers a classical objective of stream learning: compressing potentially unbounded streams into bounded-memory representations while maximizing information density. To this end, we leverage Approximate Nearest Neighbor Search (ANNS), a family of techniques that trades a small amount of retrieval accuracy for substantial gains in computational efficiency \cite{wang2021comprehensive}. Among the available approaches, graph-based methods have emerged as particularly effective due to their ability to explicitly encode neighborhood structure while requiring relatively few distance evaluations \cite{li2019approximate,fu2021high}. At the core of these methods are four foundation structures: \textit{Delaunay Graphs}, \textit{Relative Neighborhood Graphs}, \textit{Minimum Spanning Trees}, and \textit{K-Nearest Neighbor Graphs (KNNG)} \cite{wang2021comprehensive}. We adopt the latter as a directed graph with limited number of neighbors per vertex \cite{paredes2005using}.

Importantly, this data-centric perspective aligns naturally with work on hierarchical inference \cite{behera2025exploring}, where computation is dynamically allocated based on input difficulty to optimize misclassification risk and communication overhead, e.g. transmission energy \cite{nikoloska2020data}, or remote computation costs \cite{al2023case}.

\section{Methodology}

We consider a tabular data stream in which each instance $\mathbf{x}i \in \mathbb{R}^d$ is associated with a label $y_i$. Tabular Foundation Models (TFMs) perform in-context inference by conditioning their predictions on a labeled support set. Given a support set $D{\mathrm{train}}$ and a query $\mathbf{x}_q$, a TFM produces a predictive distribution over the corresponding label $y_q$:
\begin{equation}
p_\theta(y_q \mid \mathbf{x}_q, D_{\mathrm{train}}) =
\frac{\exp\big(f_\theta(\mathbf{x}_q, D_{\mathrm{train}})[y_q]\big)}
{\sum_{c=1}^{C} \exp\big(f_\theta(\mathbf{x}_q, D_{\mathrm{train}})[c]\big)}.
\end{equation}

\textbf{NNS.} However, the quadratic growth of memory and computation with context length in transformer architectures imposes a constraint on the number of support examples that can be included. To address this, prior work has explored replacing the full dataset context with localized summaries, such as $k$-means centroids \cite{feuer2023scaling}, recognizing that local information is often most informative for tabular prediction. Formally, letting $\mathrm{kNN}(\mathbf{x}_q)$ denote the $k$ nearest neighbors of $\mathbf{x}_q$ in $D_{\mathrm{train}}$, the predictive distribution becomes:
\begin{equation}
p_\theta(y \mid \mathbf{x}_q, D_{\mathrm{train}}) =
\frac{\exp\big(f_\theta(\mathbf{x}_q, \mathrm{kNN}(\mathbf{x}_q))[y]\big)}
{\sum_{c=1}^{C} \exp\big(f_\theta(\mathbf{x}_q, \mathrm{kNN}(\mathbf{x}_q))[c]\big)}.
\end{equation}

\textbf{Approximate NNS.} However, in streaming, the challenge is not merely to select a local context, but to \emph{maintain a compact memory over time} from which informative local contexts can be efficiently retrieved. In this work, we adopt approximate \textit{K-Nearest Neighbor Graphs (KNNG)} which limit the number of neighbors per vertex to $K$, producing a directed graph that is memory-efficient and suitable for large datasets \cite{paredes2005using}. For initialization, the most popular approach is the NN-Descent \cite{dong2011efficient}, which first randomly selects neighbors for each point, and then update each point’s neighbors with neighborhood propagation. As seen in algorithms such as DPG \cite{li2019approximate}, NSSG \cite{fu2021high}, and SWINN \cite{mastelini2024swinn}, NN-Descent improves neighbor distribution and search efficiency without sacrificing incremental adaptability. For \textit{HINT}, we adopt SWINN, leveraging only the NN-descent without auxiliary structures. At time $t$, a directed, weighted graph $G_t = (V_t, E_t)$ is maintained, where each vertex $v \in V_t$ corresponds to an instance in $\mathcal{W}_t$. Each vertex has at most $K$ outgoing edges, and edge weights are defined by a user-specified distance metric $d(\cdot,\cdot)$. The direct neighborhood of a vertex $v$ is defined as $N_v = \{u \in V_t \mid (v,u) \in E_t\},$ while the reverse neighborhood is $N'_v = \{u \in V_t \mid (u,v) \in E_t\}.$ The total neighborhood is given by $T_v = N_v \cup N'_v$. With this representation, there are three main sliding window operations:

\begin{itemize}
    \item \textbf{Initialization.} During an initial warm-up period, incoming instances are buffered until $|\mathcal{W}_t| = L$. During this phase, all nearest neighbor queries are answered using a linear scan. Once the window is full, an initial random directed graph is created by assigning $K$ random outgoing edges to each vertex, which is then refined iteratively.
    \item \textbf{Element insertion.} When a new instance arrives, a $K$-NN search is performed on the current graph to identify its nearest neighbors. A new vertex is then added to the graph with directed edges to these neighbors, and reverse neighborhoods are updated accordingly. The cost of insertion is equivalent to a single graph search with $k = K$.
    \item \textbf{Element removal.} When the sliding window is full, the oldest vertex is removed together with all incident edges. To preserve graph connectivity, local repairs are restrictively performed to the neighborhood of the removed vertex. Vertices that lose all incoming or outgoing connections are reconnected by performing new graph searches seeded from nearby vertices. A local refinement step is then applied to the affected neighborhood. The removal cost is $\mathcal{O}(|T_v| + \text{max}_c^2 |T_v|)$.
\end{itemize}

\begin{figure*}[t]
    \centering
    \includegraphics[width=\textwidth]{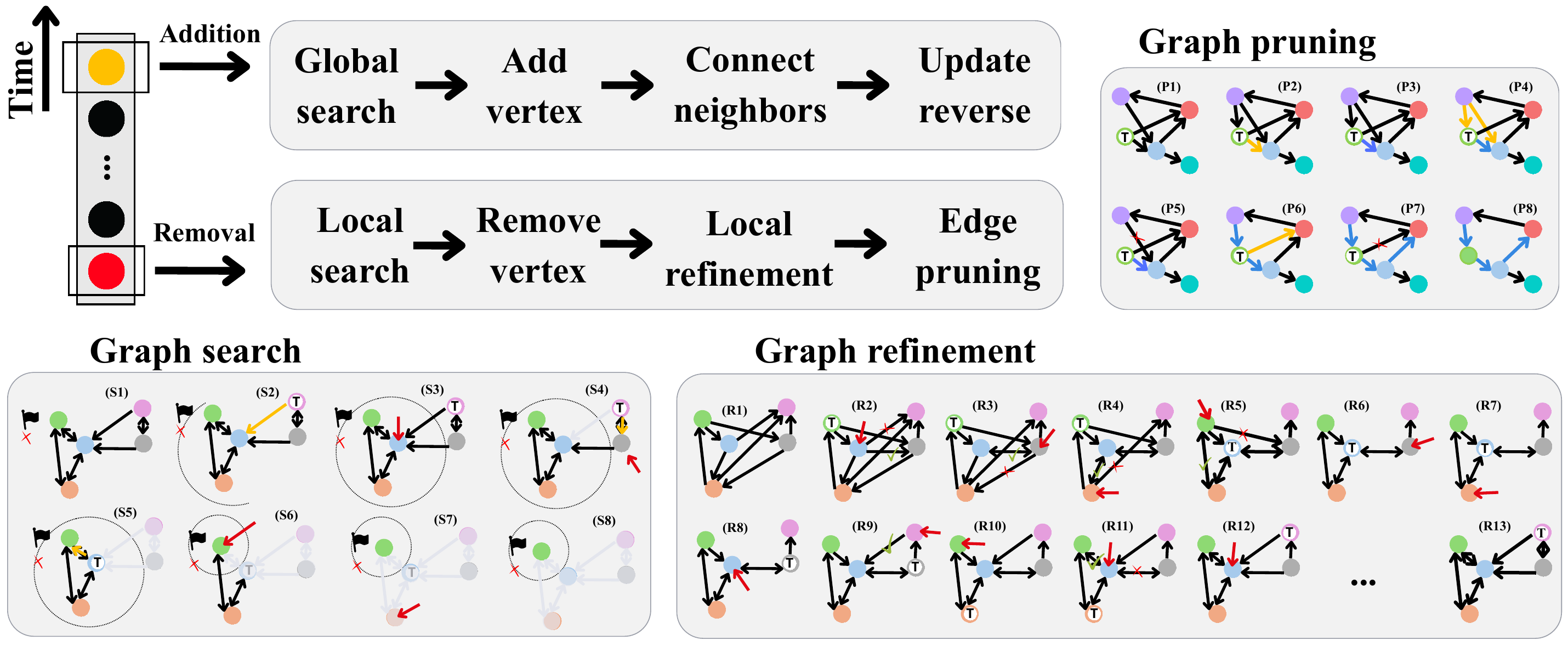}
    \caption{\textbf{Approximate addition/removal of elements in NN-based graph} relies on three main operators. \textbf{(1)} Greedy graph traversal starts from a random seed and iteratively explores closer neighbors. \textbf{(2)} Neighborhoods are iteratively refined through joins, and edge replacements. \textbf{(3)} Redundant edges are removed.}
    \label{fig:nearest}
\end{figure*}

\textbf{Graph operators.} Implementing these operations then relies on three different operators for incremental updates, as illustrated in Figure \ref{fig:nearest}:

\begin{itemize}
    \item \textbf{Search.} NN queries are answered using a greedy best-first traversal of the graph, with exploration bounded by the best-known distance, optionally relaxed by a factor $(1+\varepsilon)$ to improve recall. The search maintains a min-heap of candidate vertices to explore and a max-heap of size $k$ storing the current nearest neighbors. Vertices are explored in increasing order of distance, and candidates are added only if they improve the current distance bound defined by the $k$-th nearest neighbor. In practice, the number of graph hops remains small, yielding sublinear search time with respect to $L$.
    \item \textbf{Refinement.} Following the principle that neighbors of neighbors are likely to be neighbors, for each vertex $v$, candidate neighbors are drawn from its total neighborhood $T_v$. If $|T_v|$ exceeds a user-defined threshold $\text{max}_c$, a random subset of size $\text{max}_c$ is sampled. Pairwise neighborhood joins are then attempted among candidates to replace existing edges with closer alternatives. To avoid redundant distance evaluations, each directed edge is associated with a binary flag after participating in a join operation. Refinement proceeds iteratively and stops when either a maximum number of iterations is reached or when the number of edge updates in an iteration falls below a tolerance threshold $\delta K |V_t|$. Refinement complexity is $\mathcal{O}(\text{max}_c^2 |V_t|)$.
    \item \textbf{Pruning.} To reduce redundancy and control memory usage, a local edge pruning strategy is applied. For each vertex whose total neighborhood size exceeds $\text{max}_c$, neighbors are ordered by distance and evaluated using a triangle-based rule: in any triangle formed by vertices $(v,s,c)$, the longest edge is removed while preserving reachability. Edge removal is applied probabilistically using a parameter $\text{pruneprob} \in [0,1]$, allowing a trade-off between search speed and robustness against local minima.
\end{itemize}

\textbf{Hierarchical inference.} Leveraging this two-stage process of (1) local graph-based context retrieval, and (2) TFM inference; \textit{HINT} instantiates edge-cloud collaboration with TFMs as a hierarchical inference problem \cite{behera2025exploring}, as illustrated in Fig.~\ref{fig:vamos}, governed by: an offloading mechanism governed by a confidence threshold $\theta$, and a neighborhood-size strategy $k(t)$. Given a query point $x$, let $\mathcal{N}_k(x)$ denote its set of $k$ nearest neighbors. We define a weighted class support function for each class $j$ as
\[
s_j(x) =
\frac{\sum_{x_i \in \mathcal{N}_k(x)} \mathbb{1}(y_i = j)\, w(x, x_i)}
{\sum_{x_i \in \mathcal{N}_k(x)} w(x, x_i)},
\]
where $w(x, x_i) = d(x, x_i)^{-1}$ is a distance-based similarity weight and $d(\cdot,\cdot)$ denotes the Euclidean distance. The predicted label is given by $\hat{y}(x) = \arg\max_j s_j(x)$, and we define the confidence of the prediction as $p(x) = s_{\hat{y}(x)}(x) \in [0,1]$. The edge device decides whether to process a sample locally or offload it to the server according to the rule $\text{offload if } p(x) < \theta$, otherwise the prediction is accepted locally. To control communication overhead, the context size $k$ follows a predefined schedule $k(t) \in \mathcal{K} = {k_1,\dots,k_m}$ over the stream. Two parallel local classifiers are maintained: a fixed model using a constant small $k_{\min}$ and a scheduled model using $k(t)$ \cite{hidalgo2023paired}. Their cumulative accuracy estimates $A_t^{(c)}$ are updated online, and the classifier with the highest cumulative accuracy is selected for local prediction. For the objective function, we model the resource trade-off through a communication-oriented proxy that combines predictive errors with the number of samples transmitted to the cloud. The parameter $\beta$ controls the relative importance assigned to communication. This formulation intentionally abstracts away hardware- and deployment-specific factors such as network latency, transmission energy, memory usage, batching, and inference runtime, which depend on the underlying edge, network, and cloud infrastructure. Since TabPFN performs a fresh in-context inference for each query, we assume that the query and its selected context are transmitted for every offloaded instance, without caching or reuse across consecutive queries. The resulting cost should therefore be interpreted as a simplified communication--error proxy rather than a direct measure of end-to-end system cost. Under this formulation, the amortized cost per sample $i$ at time $t$ is:
\[
\text{Cost}_i(t) =
\begin{cases} \beta (1+k_t) + \eta_i, & \text{if } p_i < \theta \\
\gamma_i, & \text{if } p_i \geq \theta
\end{cases}
\]

where, $\eta_i$ and $\gamma_i$ indicate incorrect server and local inference errors. The formulation could in principle be extended with an online optimization procedure over a sliding window to jointly select $\theta$ and $k(t)$. We deliberately leave this optimization outside the scope of this work, as controlled parameter sweeps allow the effects of each parameter to be isolated and interpreted. Specifically, we evaluate $k(t) \in {(5,10,20,40)}$, $\theta \in {(0,0.3,0.5,0.55,0.6,0.7,0.8,0.9,1)}$ paired with a fixed reference classifier at $k_{\min}=3$. Across experiments, the edge retriever maintains a sliding window of $L=1000$ instances over a directed graph with $K=20$ neighbors per vertex.

\section{Experiments}

A prequential evaluation strategy is used, where each instance is first used to test and then update the classifier in an online manner. The streams are drawn from the USP Data Stream Repository, encompassing both binary and multiclass classification tasks. For all datasets, we consider up to the first 20k instances and include an Adaptive Random Forest (ARF) \cite{gomes2017adaptive} and a Hoeffding Tree (HT) \cite{lourencco2026dfdt} as baselines. As the TFM, we use TabPFN \cite{grinsztajn2025tabpfn}. A key practical advantage of TabPFN, and TFMs in general, is their minimal reliance on hyperparameter tuning. In TFMs, the primary source of variability arises from the ensemble prediction mechanism, in which feature orderings and scalings are randomly permuted to account for the permutation invariance of tabular columns. In our experiments, four permutations are used. Increasing the number of permutations reduces stochasticity and provides a closer approximation to the Bayesian posterior, typically improving predictive performance.

\begin{figure}[h]
    \centering
    \includegraphics[width=0.66\linewidth]{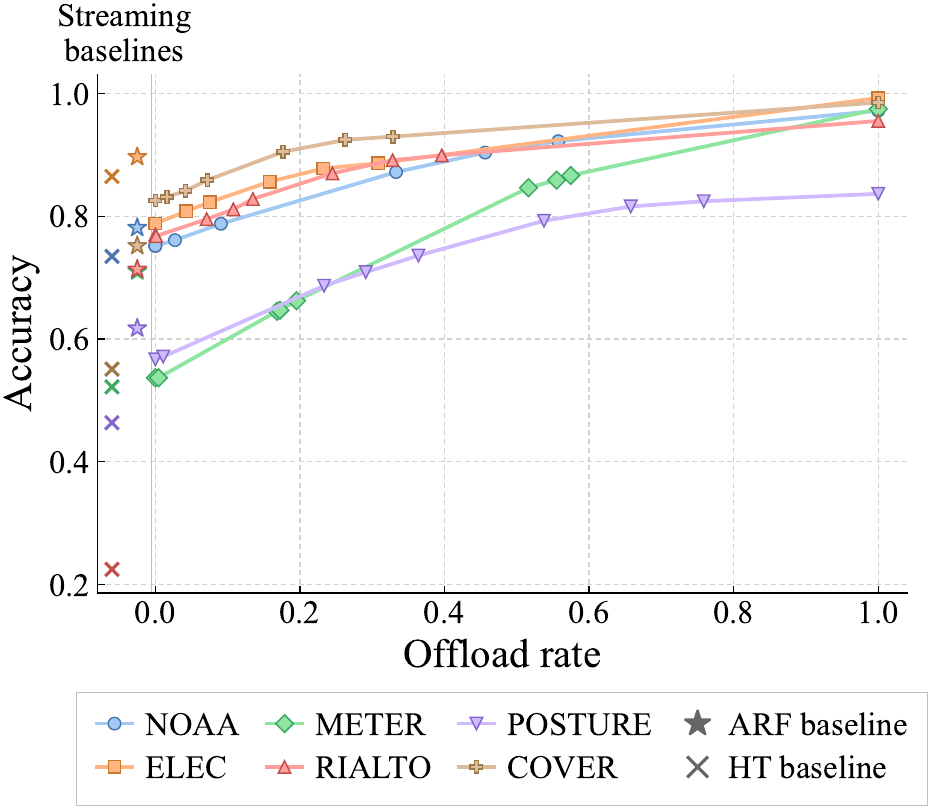}
    \caption{Offload effect on \textit{HINT}'s accuracy.}
    \label{fig:overview}
\end{figure}

\textbf{Effect of $\theta$ on accuracy.} Figure~\ref{fig:overview} illustrates the effect of the offload rate, while Table~\ref{tab:main} summarizes accuracy and balanced accuracy across baselines and hierarchical inference (HI) at $\theta=0.7$. HI outperforms ARF on five of six datasets (all except ELEC), while offloading only 16--54\% of the stream. Compared with local-only inference, HI provides substantial accuracy improvements across all six datasets. The gains are most pronounced under class imbalance: on POSTURE, ARF collapses to majority-class predictions (BAC 0.285), whereas HI achieves 0.555 BAC. The local-only strategy remains competitive on RIALTO and COVER.

\begin{table}[t]
\centering
\setlength{\tabcolsep}{4pt}
\small
\caption{Hierarchical inference at the working point $\theta=0.7$ with its offload rate. Best Accuracy and Balanced Accuracy in \textbf{bold}; full offloading shown as a reference upper bound. Datasets are clipped to the first 20k instances.}
\label{tab:main}
\begin{tabular}{lccccrc}
\toprule
Dataset & ARF & HT & Local-only & HI ($\theta{=}0.7$) & Off. & Full-offload \\
 & {\footnotesize ACC/BAC} & {\footnotesize ACC/BAC} & {\footnotesize ACC/BAC} & {\footnotesize ACC/BAC} & rate & {\footnotesize ACC/BAC} \\
\midrule
NOAA    & 78.1/70.8 & 73.5/68.3 & 75.2/69.0 & \textbf{87.2/83.2} & 33.0 & 97.2/96.5 \\
ELEC    & \textbf{89.7/89.3} & 86.5/86.2 & 78.8/78.4 & 85.7/85.4 & 16.0 & 99.3/99.2 \\
METER   & 71.1/71.1 & 52.2/52.2 & 53.7/53.7 & \textbf{84.7/84.7} & 52.0 & 97.5/97.5 \\
RIALTO  & 71.4/71.4 & 22.5/22.5 & 77.0/77.0 & \textbf{87.0/87.0} & 24.0 & 95.5/95.5 \\
POSTURE & 61.8/28.5 & 46.4/21.7 & 56.7/28.4 & \textbf{79.3/55.5} & 54.0 & 83.7/61.0 \\
COVER   & 75.2/74.1 & 55.1/49.8 & 82.6/82.6 & \textbf{90.5/90.7} & 18.0 & 98.5/98.5 \\
\midrule
\textbf{Average} & 74.6/67.5 & 56.0/50.1 & 70.7/64.9 &
\textbf{85.7/81.1} & 32.8 & 95.3/91.4 \\
\bottomrule
\end{tabular}
\end{table}

\begin{figure}[h]
    \centering
    \includegraphics[width=\textwidth]{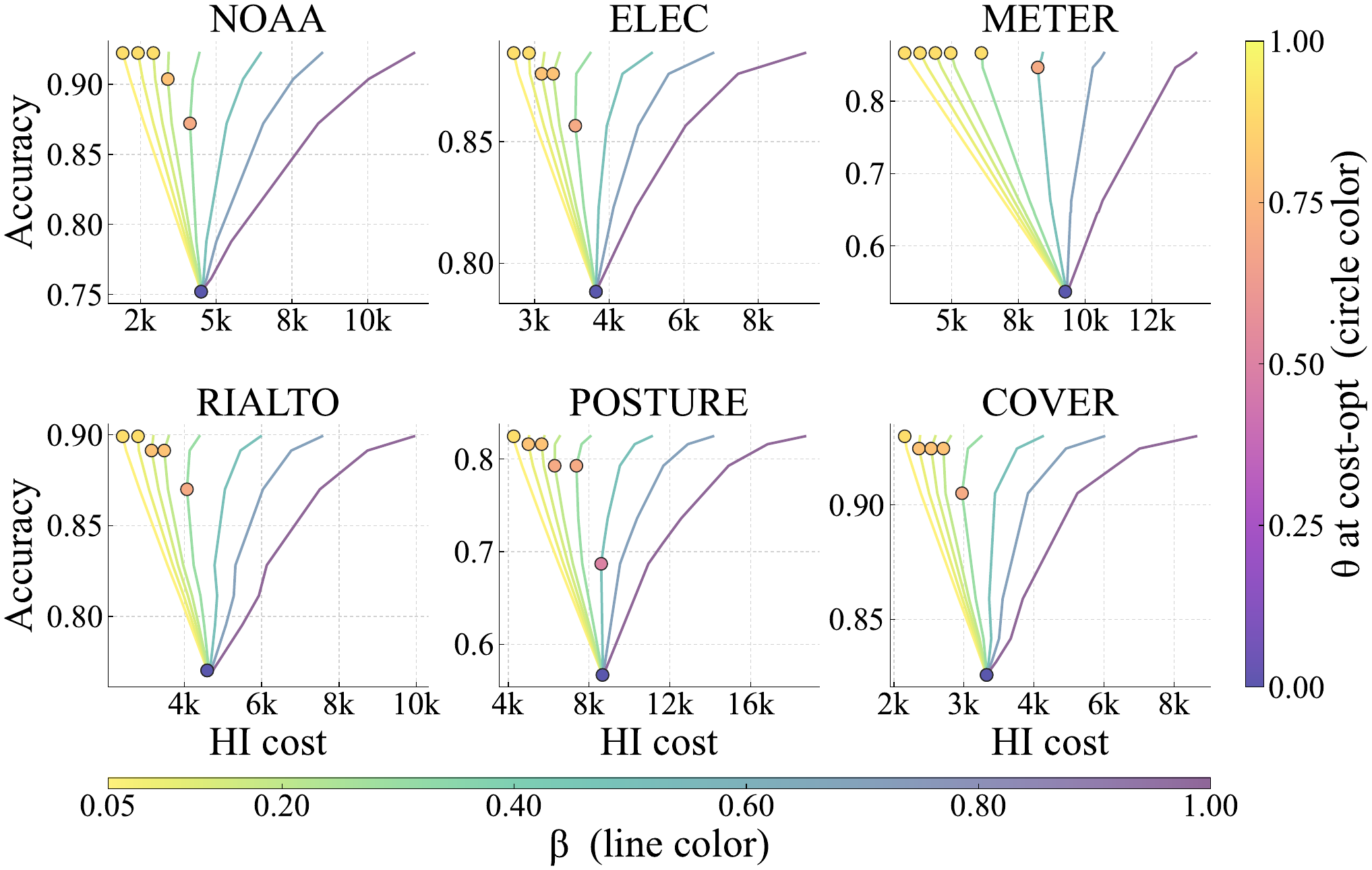}
        \caption{Hierarchical inference cost x Accuracy.}
    \label{fig:overview2}
\end{figure}

\textbf{Effect of $\theta$ on HI cost.} Figure \ref{fig:overview2} illustrates the trade-off between inference accuracy and HI cost across the six datasets under different cost sensitivity. Each curve corresponds to a distinct value of $\beta$, while each point represents an operating configuration obtained from the controlled parameter sweep over $\theta$ and $k(t)$. Across all datasets, HI induces a consistent and well-defined Pareto frontier. In all cases, accuracy increases rapidly at low to moderate cost regimes, followed by a clear saturation effect where further cost increases yield diminishing improvements in performance. This convex structure indicates that near-optimal accuracy can be achieved without resorting to full offloading, highlighting the efficiency of hierarchical inference as a mechanism for reducing communication overhead while preserving predictive quality. The cost sensitivity parameter $\beta$ systematically controls the shape and position of the Pareto frontier. Lower values of $\beta$ result in policies that favor more frequent offloading, yielding higher accuracy at moderate cost. In contrast, higher $\beta$ values penalize communication more strongly, shifting the operating points toward lower-cost regimes with reduced accuracy. For example, for METER, a full-offload strategy attains an accuracy of 97.47\% at a cost of $19{,}995\beta + 506$, whereas a no-offload (local-only) strategy yields 53.69\% accuracy at a cost of $9{,}259$ (all incurred as local misclassifications). Within the evaluated configurations, the lowest-cost operating point for $\beta=0.5$ corresponds to $\theta^*=0.7$, under which 51.6\% of samples are offloaded (10{,}320 instances) and 48.4\% are processed locally (9{,}675 instances). This configuration yields an accuracy of 84.65\% at a cost of $10{,}320\beta + 3{,}069$. Relative to full offloading, HI reduces the offloaded communication volume by 48.4\% while retaining 86.8\% of its accuracy (84.65\% vs.~97.47\%) and substantially exceeding the local-only baseline (53.69\%). In terms of the communication--error proxy, HI becomes cheaper than full offloading once $\beta \gtrsim 0.27$, reaching up to a 35\% total-cost reduction at $\beta=1$.

\begin{figure}[h]
    \centering
    \includegraphics[width=\textwidth]{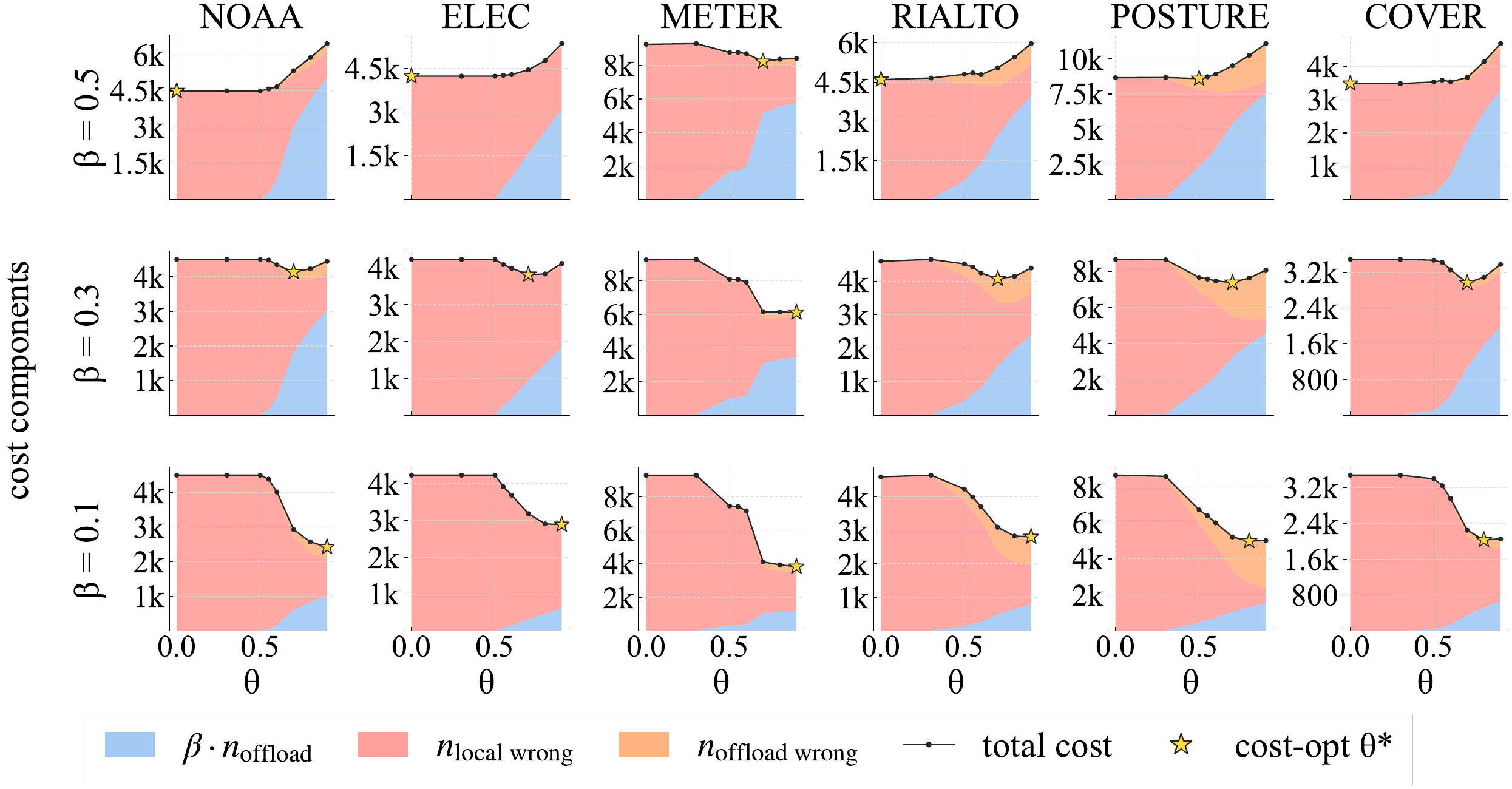}
        \caption{$\theta$ x Hierarchical inference cost comparison across different $\beta$ values.}
    \label{fig:theta_accuracy_beta}
\end{figure}

\textbf{HI cost decomposition.} Figure~\ref{fig:theta_accuracy_beta} further highlights the HI cost decomposition. For $\beta \in [0.5,1)$, NOAA, ELEC, RIALTO, and COVER favor lower offloading rates, whereas for $\beta \in [0,0.1)$, full offloading provides the lowest evaluated objective for NOAA, ELEC, and METER. COVER and ELEC achieve consistently high accuracy under HI, with relatively smooth trade-offs, while NOAA and RIALTO exhibit intermediate behavior. METER shows the highest sensitivity to cost changes, with sharper transitions between operating regimes. Nonetheless, a consistent observation across all datasets is the existence of a sweet-spot region along the Pareto frontier. In this region, the system achieves near-maximum accuracy (typically within 2--5\% of the best observed performance) while reducing the communication-oriented cost relative to high-offloading configurations through moderate values of $\theta$. POSTURE is the only dataset where a middle-ground configuration remains consistently beneficial for $\beta \in [0,0.1)$, indicating a more challenging inference setting.

\begin{figure}[h]
    \centering
    \includegraphics[width=0.85\textwidth]{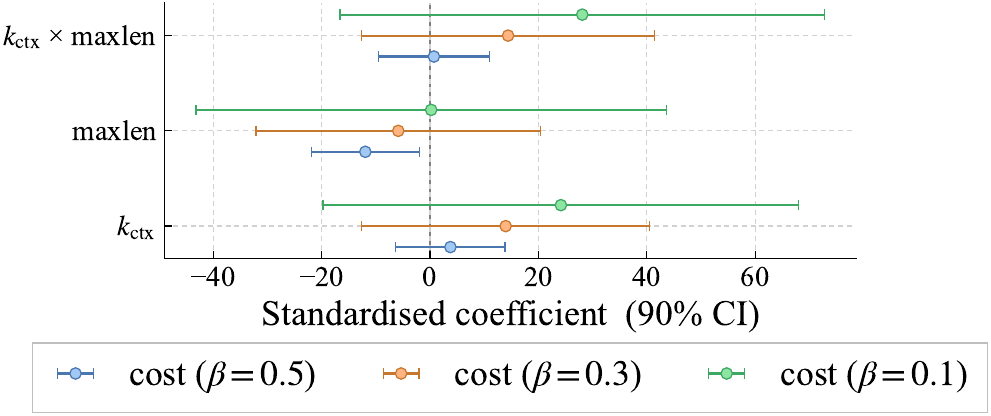}
        \caption{$k(t)$ x Hierarchical inference cost comparison across different $\beta$ values.}
    \label{fig:theta_accuracy_beta2}
\end{figure}

\textbf{Effect of $k(t)$ on HI cost.} Finally, Figure~\ref{fig:theta_accuracy_beta2} presents a controlled sweep of the neighborhood size $k(t)$ and its effect on HI cost. Across all values of $\beta$, increasing $k(t)$ consistently increases communication cost, since more samples must be retrieved and aggregated from memory. To better understand this behavior, we ablate the influence of the memory window size. In terms of accuracy, the effect of $k(t)$ is not uniform and depends on the effective richness of the memory window. For small memory windows, increasing $k(t)$ is expected to provide limited additional benefit, as the available neighborhood structure is constrained. In this regime, the additional communication cost of larger $k(t)$ is not compensated by substantial gains in predictive performance, making smaller values of $k(t)$ more efficient overall. However, Figure \ref{fig:theta_accuracy_beta2} reveals a more nuanced effect. We observe that larger windows increase the marginal utility of retrieving additional neighbors when communication costs are low ($\beta = 0.1$), and decrease it when communication costs are high ($\beta = 0.5$). Thus, increasing $k(t)$ alone does not substantially improve local accuracy, even in the presence of richer memory windows; however, it can be crucial in making the communication of a larger number of samples more beneficial. In other words, in settings where larger or more diverse memory buffers are required, such as those exhibiting recurrent patterns or concept drift, higher values of $k(t)$ may not be efficiently leveraged locally, but can be effectively exploited by the TFM in the cloud.

\section{Conclusion}

Motivated by the increasing gap between the capabilities of modern TFMs and the practical constraints imposed by streaming environments, this work introduced \textit{HINT}, a hierarchical inference framework that couples edge-based retrieval with cloud-based TFMs. We formulated a resource-aware inference mechanism that controls (i) whether a sample should be processed locally or offloaded to a TFM, and (ii) how much contextual information should be retrieved through the neighborhood size $k(t)$. Across multiple benchmark datasets, hierarchical inference induced a consistent and well-defined Pareto frontier between accuracy and communication cost under controlled parameter sweeps. This opens a promising direction for future work on online memory-aware inference policies, as well as more principled optimization strategies for joint control of communication, memory, and inference in streaming settings.

\begin{credits}
\subsubsection{\ackname} Funded by FCT under Ph.D. PRT/BD/03225/2025, and \newline PRT/BD/154713/2023, and project 10.54499/UID/00760/2025.

\subsubsection{\discintname} The authors have no competing interests to declare that are relevant to the content of this article.
\end{credits}

\bibliographystyle{splncs04}
\bibliography{references}

\end{document}